\documentclass[sigconf]{acmart}

\usepackage{booktabs}
\usepackage{float}
\usepackage{tabularx}
\AtBeginDocument{%
  }

\setcopyright{acmlicensed}
\copyrightyear{2026}
\acmYear{2026}
\acmDOI{XXXXXXX.XXXXXXX}
\acmConference[AI\&OG @ ICAIL]{First Workshop on Artificial Intelligence \& Open Government at the 21st International Conference on Artificial Intelligence and Law}{June 8, 2026}{Singapore}
\acmBooktitle{The First Workshop on Artificial Intelligence \& Open Government at the 21st International Conference on Artificial Intelligence and Law, June 8, 2026, Singapore}
\begin{document}

%%
%% The "title" command has an optional parameter,
%% allowing the author to define a "short title" to be used in page headers.
\title{To Redact, or not to Redact? A Local LLM Approach to Deliberative Process Privilege Classification}

%%
%% The "author" command and its associated commands are used to define
%% the authors and their affiliations.
%% Of note is the shared affiliation of the first two authors, and the
%% "authornote" and "authornotemark" commands
%% used to denote shared contribution to the research.
\author{Maik Larooij}
\email{m.k.larooij@uva.nl}
% \orcid{1234-5678-9012}
% \author{G.K.M. Tobin}
% \authornotemark[1]
% \email{webmaster@marysville-ohio.com}
\affiliation{%
  \institution{University of Amsterdam}
  \city{Amsterdam}
  % \state{Ohio}
  \country{The Netherlands}
}
\author{David Graus}
\email{d.p.graus@uva.nl}
\affiliation{
\institution{University of Amsterdam}
\city{Amsterdam}
\country{The Netherlands}
}
%%
%% By default, the full list of authors will be used in the page
%% headers. Often, this list is too long, and will overlap
%% other information printed in the page headers. This command allows
%% the author to define a more concise list
%% of authors' names for this purpose.
%\renewcommand{\shortauthors}{Anonymous}

%%
%% The abstract is a short summary of the work to be presented in the
%% article.
\begin{abstract}
    Government transparency laws, like the Freedom of Information (FOIA) acts in the United States and United Kingdom, and the Woo (Open Government Act) in the Netherlands, grant citizens the right to directly request documents from the government. As these documents might contain sensitive information, such as personal information or threats to national security, the laws allow governments to redact sensitive parts of the documents prior to release. 
    We build on prior research to perform automatic sensitivity classification for the FOIA Exemption 5 deliberative process privilege using Large Language Models (LLMs). 
    However, processing documents not yet cleared for review via third-party cloud APIs is often legally or politically untenable. 
    Therefore, in this work, we perform sensitivity classification with a small, local model, deployable on consumer-grade hardware (Qwen3.5 9B). 
    We compare eight variants of applying LLMs for sentence classification, using well-known prompting techniques, and find that a combination of Chain-of-Thought prompting and few-shot prompting with error-based examples outperforms classification models of earlier work in terms of recall and F2 score. 
    This method also closely approaches the performance of a widely-used, cost-efficient commercial model (Gemini 2.5 Flash). 
    In an additional analysis, we find that sentences that are predicted as deliberative contain more verbs that indicate the expression of opinions, and are more often phrased in in first-person. 
    Above all, deliberativeness seems characterized by the presence of a combination of multiple indicators, in particular the combination of first-person words with a verb for expressing opinion.
\end{abstract}

%%
%% The code below is generated by the tool at http://dl.acm.org/ccs.cfm.
%% Please copy and paste the code instead of the example below.
%%


%%
%% Keywords. The author(s) should pick words that accurately describe
%% the work being presented. Separate the keywords with commas.
%% A "teaser" image appears between the author and affiliation
%% information and the body of the document, and typically spans the
%% page.

% \received{20 February 2007}
% \received[revised]{12 March 2009}
% \received[accepted]{5 June 2009}

%%
%% This command processes the author and affiliation and title
%% information and builds the first part of the formatted document.
\maketitle

\section{Introduction}
Several countries have implemented laws that grant citizens the right to request and obtain information from the government in the form of documents. 
Examples include the Freedom of Information Acts (FOIA) in the United States and United Kingdom, and the `Wet open overheid' (Woo, Open Government Act) in the Netherlands. 
Due to the sensitive nature of these documents, the laws allow government agencies to redact parts of the documents prior to releasing them, for example when they contain personal information, threats to national security, or trade secrets. 
This requires officials to manually review all documents before publication, which is a time-consuming and error-prone task. 
The process of redacting sensitive information was identified as one of the main drivers behind substantial delays in response time \cite{grunewald1998foia, wiemers2026zwaluw}. 

Prior research has already focused on automatic classification of FOIA exemption grounds, for example on personal information (FOIA UK Section 40) and international relations (FOIA UK Section 27) \cite{mcdonald2015using, berardi2015semi, mcdonald2017enhancing, narvala2021reldiff, mcdonald2017study}, and the deliberative process privilege (FOIA US Exemption 5) \cite{baron2022providing}. 
These papers rely on training classifiers (such as Logistic Regression, SVM or BERT) for sentence classification. 
And while these papers show solid results, there are also some downsides to these methods. 
For example, the results in \cite{baron2022providing} show that performance fluctuates when the model is trained on documents of a different custodian of records or reviewer, suggesting that the models are sensitive to their training data and do not generalize well to out-of-distribution data, a common scenario in real-world cases. 
Furthermore, while sensitivity classification improves sensitivity reviewing speed and accuracy \cite{mcdonald2020accuracy}, providing useful explanations to the predictions, which the proposed models do not support, would further improve this process. 

Large Language Models (LLMs) have the potential to address these shortcomings. 
LLMs can make `out-of-the-box' predictions, making the training phase obsolete and therefore allowing to avoid the generalization problems. 
Also, LLMs are able to generate plausible (but not necessarily correct!) explanations about predictions. 
For this reason, \citet{baron2023using} already leveraged GPT 3.5 for predicting whether certain paragraphs of text should be exempt from public release based on FOIA Exemption 5 (deliberative process privilege). 
The LLM did not perform better than traditional text classification methods, but the explanations seemed `not unreasonable' and `helpful'.

We build on the research of \citet{baron2023using}, and subsequent work of \citet{branting2025decision} who augmented the dataset of \citet{baron2023using} with sentence-level annotations, in three ways: 

\begin{enumerate}
    \item We test several well-established prompting techniques (Chain-of-Thought, few-shot, multi-agent), to see whether they improve classification performance over simple zero-shot prompting;
    \item We compare the performance of a commercial, closed-source model (Gemini 2.5 Flash) to a smaller, local open-weights model (Qwen3.5-9B);
    \item We analyze the presence of syntactical indicators, like verbs for expressing opinions, first-person words and future temporal words in sentences that are unanimously predicted deliberative or non-deliberative by all prompting variants.
\end{enumerate}

The second contribution is of particular importance in the open government domain: governments typically want to process documents on their own infrastructures, especially when documents may contain sensitive information. 
This hinders the use of larger, closed commercial models. 
In terms of evaluation, we argue that recall is more important than precision, as the impact of 'forgetting' to redact sensitive information can be much larger than suggesting to redact too much. 

\section{Background}
\subsection{Deliberative Process Privilege (FOIA US Exemption 5}
\label{sec:deliberativeprocess}
The U.S. Code of Federal Regulations, Title 32, § 1662.22\footnote{From the code of federal regulations: \url{https://www.ecfr.gov/current/title-32/subtitle-B/chapter-XVI/part-1662/section-1662.22}} provides a clear explanation of deliberative process privilege: 
\textit{"This privilege protects the decision-making processes of government agencies. Information is protected under this privilege if it is predecisional and deliberative. The purpose of the privilege is to prevent injury to the quality of the agency decision-making process by encouraging open and frank internal discussions, by avoiding premature disclosure of decisions not yet adopted, and by avoiding the public confusion that might result from disclosing reasons that were not in fact the ultimate grounds for an agency's decision."} 

From this explanation we infer that, for a record to be protected by this privilege, it must be both \emph{predecisional} and \emph{deliberative}. 
A record is \emph{predecisional} if it is created before a final decision to assist a decision-maker.
Its primary characteristic is that it precedes an agency's action. 
It excludes "final opinions" or documents that justify and explain a decision that has already been made or a matter that has already been officially settled. 
A record is \emph{deliberative} if it reflects internal decision-making processes, such as personal recommendations, legal advice, or policy opinions. 
The content must be the personal view of the author, rather than an already established policy of the agency.
Examples include draft document or documents that are recommendatory in nature.

While not exactly the same, other jurisdictions know similar exemption grounds. 
In the UK, Section 35 protects Ministerial communications and the formulation or development of government policy. 
Article 5.2 of the Dutch Woo protects deliberative policy views held by civil servants in the course of policy formulation. 

\subsection{Sensitivity Classification and Applications}
There have already been numerous efforts in predicting whether text should be exempt according to Freedom of Information laws. 
\citet{mcdonald2014towards} showed the first SVM classifier to classify whether exemptions (Section 27: International Relations, Section 40: Personal Information) from the UK Freedom of Information Act 2000 apply to textual documents. 
Later work builds upon the same (non-public) dataset, by extending the classifier with grammatical features (Part-of-Speech N-grams) \cite{mcdonald2015using}, semantic features (embeddings) \cite{mcdonald2017enhancing}, different SVM kernel functions \cite{mcdonald2017study}, and knowledge graph relation representations \cite{narvala2021reldiff}.

Another, more recent line of research on the US Freedom of Information Act focuses on the more complex challenge of classifying whether a paragraph of text falls under Exemption 5's deliberative process privilege. 
\citet{baron2022providing} compare Logistic Regression (LR), SVMs, a Begin-Inside-Outside (BIO)-tagger, and keyword search, and find the BIO variant performs best. 
In addition, using ChatGPT for the deliberative process privilege showed promising results in terms of explainability, but did not perform better than traditional text classification methods. 
The work by \citet{branting2025decision} used new annotations at the sentence-level, instead of the paragraph-level, and also evaluated LR, SVMs, and BERT models. 

Augmenting human sensitivity review with automatic sensitivity classification has been shown to substantially improve classification accuracy and reviewing speed \cite{mcdonald2020accuracy}. 
Sensitivity classification is also combined with active learning, where the classifier is improved by hand-labeling documents that will likely provide the most valuable evidence \cite{mcdonald2018active}. 
More recently, clustering documents by latent semantic categories or coherent document groups in combination with active learning increased the efficiency of human reviewers \cite{narvala2022role, narvala2022sensitivity}.

Sensitivity classification is also used to directly search among sensitive content. 
\citet{oard2016evaluating} introduced the "Cost-Sensitive Discounted Cumulative Gain" to jointly optimize reducing sensitivity, and maximizing relevance in search results \cite{sayed2019jointly, mckechnie2024bi}. 
Sensitivity classification is used here to approximate the sensitivity of search results. 
It has also been proposed in combination with sensitivity-aware pseudo-relevance feedback approaches \cite{mckechnie2024cascading}. 

\section{Experiments}

\subsection{Dataset}
We leverage the newly annotated corpus\footnote{The full dataset can be found at: \url{https://github.com/nater82/ChatGPT\_FOIA\_Exemption5\_Data}} of \citet{branting2025decision}, which is an adaptation of the corpus collected by \citet{baron2022providing}. 
The corpus contains five batches of Clinton White House documents and e-mails of two White House custodians: Elena Kegan (batches K1, K2, K3, K5) and Cythia Rice (batch R4). 
The documents were split to the sentence level, with label AD (always deliberative, regardless of context) or Non-AD. 

For clarity, we paraphrase two examples mentioned in the paper:

\begin{enumerate}
    \item AD: \emph{“You could also announce that you will expand AmeriCorps to include a new child-care corps.”}
    \item Non-AD: \emph{“A quick survey of the programs identified above indicates that up to 20,000 prisoners may be receiving benefits improperly.}
\end{enumerate}

Like the original authors, we exclude batch K5 due to the comparatively large class imbalance (8\% AD). 
For the remaining batches, we show the size and class balance (\% AD-labeled) in Table~\ref{tab:dataset}. 

\begin{table}[h]
    \centering
    \begin{tabular}{crr}
        \toprule
        \textbf{Batch} & \# \textbf{Sentences} & \textbf{AD}  \\
        \midrule
        K1 & 1,262 & 270 (21\%) \\
        K2 & 992 & 411 (45\%) \\
        K3 & 1,522 & 400 (26\%) \\
        R4 & 728 & 210 (29\%) \\
        \midrule
        \textbf{Total} & \textbf{4,502} & \textbf{1,291 (29\%)} \\
        \bottomrule
    \end{tabular}
    \caption{Batch volumes and number of deliberative items.}
    \label{tab:dataset}
\end{table}

\subsection{Models}
We compare two LLMs with different characteristics, representing real-world deployment scenarios where both operational costs and data-sovereignty and security constraints affect model selection, particularly in the sensitive context of automated FOIA classification.

Therefore we select two models:
\begin{enumerate}
    \item We select a widely-used, cost-efficient \textbf{commercial} model: Gemini 2.5 Flash, which is explicitly positioned for high-volume, latency- and cost-sensitive workflows, making it a realistic choice for what a governmental agency would deploy with high FOIA volumes. The model is optimized for high throughput and low latency, making it a proxy for a production-grade LLM that potentially needs to process thousands of pages of text in short time, while costs need to be kept low.
    \item As processing documents not yet cleared for review via third-party cloud APIs is often legally or politically untenable, we pick an open-weights model that is small enough for \textbf{local} (air-gapped) deployment on consumer-grade hardware. 
    The Qwen3.5 9B model is, at the time of writing, one of the best small and open models available for on-premise deployment. The 9B parameters model is small enough to run on consumer-grade GPUs yet it rivals larger models in performance \cite{qwen3.5}.
\end{enumerate}

For the Gemini model, we leveraged Google's Gemini API service. 
We ran Qwen3.5 9B on a single 24GB NVIDIA A10 GPU. 
For both models, we set the temperature to 0 to minimize output variability. 

The Logistic Regression, SVM and BERT models used by \citet{branting2025decision} serve as a baseline (without rerunning) for our LLM-based sensitivity classification. 

\begin{table*}[!ht]
    \centering
    \caption{Classification results for the combined set of sentences (N=4,502).
    Best results per variant group (\citet{branting2025decision}, Qwen3.5 9B, Gemini 2.5 Flash) are marked in bold. 
    All few-shot experiments use $k=5$ examples.}
    \label{tab:table1}
    \small
    \begin{tabular}{lccccc}
    \toprule
    \textbf{Variant} & \textbf{Prec} & \textbf{Rec} & \textbf{F1} & \textbf{F2} & \textbf{MCC} \\
    \midrule
    \multicolumn{6}{l}{\textit{\citet{branting2025decision}}} \\
    \midrule
    LR (10-fold CV)   & 0.620 & \textbf{0.720} & 0.666 & 0.697 & 0.490 \\
    SVM (10-fold CV)  & 0.680 & 0.590 & 0.632 & 0.606 & 0.480 \\
    BERT (10-fold CV) & \textbf{0.730} & 0.710 & \textbf{0.720} & \textbf{0.714} & \textbf{0.590} \\
    \midrule
    \multicolumn{6}{l}{\textit{Qwen3.5 9B}} \\
    \midrule
    Zero-Shot (baseline)                                & 0.708          & 0.458          & 0.556          & 0.493          & 0.444          \\
    Few-Shot                                         & 0.580          & 0.603          & 0.591          & 0.599          & 0.423          \\
    Few-Shot (Error-based)                           & 0.646          & 0.615          & 0.630          & 0.621          & \textbf{0.486} \\
    CoT                                              & 0.542          & 0.746          & 0.628          & 0.694          & 0.455          \\
    CoT + Few-Shot (Error-based)                     & 0.542          & \textbf{0.800} & \textbf{0.646} & \textbf{0.731} & 0.484          \\
    Multi-agent                                      & \textbf{0.769} & 0.400          & 0.526          & 0.442          & 0.446          \\
    CoT + Multi-agent                                & 0.539          & 0.477          & 0.506          & 0.488          & 0.326          \\
    CoT + Few-Shot (Error-based) + Multi-agent       & 0.561          & 0.456          & 0.503          & 0.474          & 0.334          \\
    \midrule
    \multicolumn{6}{l}{\textit{Gemini 2.5 Flash}} \\
    \midrule
    Zero-Shot (baseline)                                & 0.554          & \textbf{0.808} & \textbf{0.658} & \textbf{0.740} & \textbf{0.502} \\
    Few-Shot                                         & 0.581          & 0.654          & 0.615          & 0.638          & 0.449          \\
    Few-Shot (Error-based)                           & \textbf{0.652} & 0.637          & 0.644          & 0.640          & \textbf{0.504} \\
    CoT                                              & 0.507          & 0.778          & 0.614          & 0.703          & 0.432          \\
    CoT + Few-Shot (Error-based)                     & 0.558          & 0.760          & 0.643          & 0.709          & 0.480          \\
    Multi-agent                                      & 0.564          & 0.787          & \textbf{0.657} & 0.729          & 0.501          \\
    CoT + Multi-agent                                & 0.540          & 0.738          & 0.623          & 0.688          & 0.449          \\
    CoT + Few-Shot (Error-based) + Multi-agent       & 0.553          & 0.734          & 0.631          & 0.689          & 0.462          \\
    \bottomrule
    \end{tabular}
\end{table*}

\subsection{Variants}

We test eight different prompting techniques, which are explained below. 
In essence, all variants have the same output: predict whether a sentence of text is AD (always deliberative), with the output being a JSON of the form \texttt{\{..., "deliberative": 0 or 1\}}. 

\subsubsection{Zero-Shot (baseline)}
This is the prompt in its simplest form: \emph{"Would the following be considered deliberative under FOIA exemption 5?"} 
Note that this forces the LLM to use its parametric knowledge to resolve what the deliberative process privilege is. 
This is also (almost) one of the versions used by \citet{baron2023using}.

\subsubsection{Few-shot prompting}
A commonly used technique is to provide the LLM with $k$ example classifications, which can be both AD and non-AD. 
We chose to include 5 examples in the prompt (i.e., $k=5$); our goal is not to find the most optimal number of examples, but rather evaluate the technique as a whole. 
A single example looks like this: \texttt{"Sentence: "You could also announce that you will expand AmeriCorps to include a new child-care corps." Correct label: ALWAYS DELIBERATIVE (deliberative=1)"}.
Examples are randomly selected (we use a seed for evaluation purposes) from all the other batches, so they are never sampled from the batch in process. 

\subsubsection{Error-based Few-shot prompting}
This configuration is the same as the few-shot variant described above, however, the examples are no longer randomly drawn from the remaining batches, but rather from a subset of sentences that the zero-shot model classified incorrectly (still only selected from other batches, never from the batch under consideration). 
Using these "harder" sentences as examples may help to correctly classify more challenging cases.

\subsubsection{Chain-of-Thought prompting (CoT)}
Chain-of-Thought (CoT) prompting is a technique that forces the LLM to reason about the choice it makes \cite{wei2022chain}. 
We divide the process of determining whether a sentence is always deliberative in three steps, and append that to the system prompt:
\begin{enumerate}
    \item Predecisional: Does this sentence concern an issue where a final decision has not yet been made?
    \item Deliberative: Does it express an opinion, recommendation, or internal deliberation rather than a fact?
    \item Conclude: Based on steps 1 and 2, is this sentence always deliberative? 
\end{enumerate}

Additionally, we provide the information mentioned in section \ref{sec:deliberativeprocess} about when a record is considered predecisional and deliberative. 
The LLM is forced to output in this format to reason about the steps: \texttt{\{"step1": "...", "step2": "...", "deliberative": 0 or 1\}}.

\subsubsection{CoT + Few-shot (hard)}
This is a combination of our CoT and error-based few-shot prompting variants. 
However, we now also generate 'correct' reasoning steps and include those in the examples. 

\subsubsection{Multi-Agent (Majority Vote)}
We leverage a three-agent setup: 
the first agent predicts whether the sentence is always deliberative (same approach as our zero-shot baseline), 
the second agent is shown the sentence and the first agent's prediction, and makes a prediction of its own. 
When the first two agents agree, this is considered the final prediction. 
If there is disagreement, a third agent is shown both predictions, the sentence, and makes the tie-breaking vote.

\subsubsection{CoT + Multi-Agent (Predictor-Critic-Judge)}
This is a combination of the CoT and multi-agent variants. 
However, the agent roles are now made more explicit with system prompts:

\begin{enumerate}
    \item Agent 1 (Predictor): makes an initial prediction with CoT reasoning;
    \item Agent 2 (Critic): reviews the predictor's reasoning and flags and explains any flaws in the reasoning, also gives a new suggestion (output: \texttt{\{"assessment": "sound" or "flawed", "issues": "describe any issues, or 'none' if sound", "suggestion": 0 or 1\}})
    \item Agent 3 (Judge): reviews both the initial prediction and the critical review and makes an explained final decision. Only comes into play when the predictor and critic disagree (output: \texttt{\{"rationale": "...", "deliberative": 0 or 1\}}).
\end{enumerate}

\subsubsection{CoT + Few-shot (hard) + Multi-Agent}
This is a combination of CoT, error-based few-shot prompting and the multi-agent setup. 
The only difference with the variant directly above is that the first agent (predictor) is now supplied with five error-based examples in the prompt.

\subsection{Evaluation}
Just like \citet{branting2025decision} we report precision, recall and MCC (Matthew's Correlation Coefficient). 
For completeness, we also report the F1 score, although we argue that recall is more important than precision, as the impact of `forgetting’ to redact a piece of sensitive information is larger than suggesting to redact too much. 
To this end, we also leverage the F2 score that weighs recall twice as much as precision, and use recall and F2 score as the leading metrics in the results section. 

We compare performance on the combined set of sentences (N=4,502) from of all batches (K1, K2, K3, R4) to the 10-fold cross-validation LR, SVM and BERT results by \citet{branting2025decision}. 

As we saw in earlier work \cite{baron2022providing}, classification performance fluctuates when the model is trained on documents of a different custodian of records or reviewer.
Moreover, results from \cite{branting2025decision} show that the difficulty of predicting deliberativeness differs per batch (best recall 0.6 on hardest batch K2, 0.78 on easiest batch K1).
In the table, we compare our variants to LR, SVM and BERT models as reported by \citet{branting2025decision}.
The authors applied 10-fold cross-validation on "the union of sentences irrespective of batch boundaries or document boundaries within batches". 
This risks pooling training data of the same custodian used for testing, and does not take into account the varying custodians, reviewers and difficulty of batches. 
For this reason, we also report precision and recall on the hardest batch K2 (in terms of recall).

The prompts and code used for this paper is available on GitHub\footnote{\url{https://github.com/opengov-lab/to-redact-or-not-to-redact}}.

\begin{figure*}[!h]
    \centering
    \includegraphics[width=0.9\linewidth]{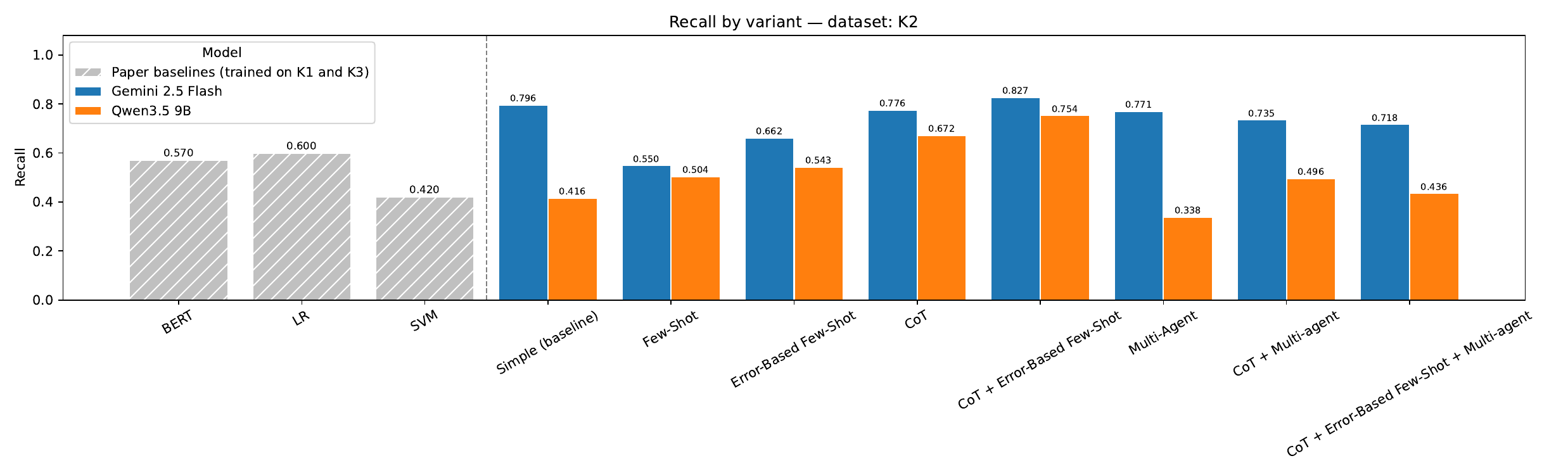}
    \caption{Comparison of recall of deliberative sentences (AD) in batch K2 for the different models and variants.}
    \label{fig:fig1}
\end{figure*}
\begin{figure*}
    \centering
    \includegraphics[width=0.9\linewidth]{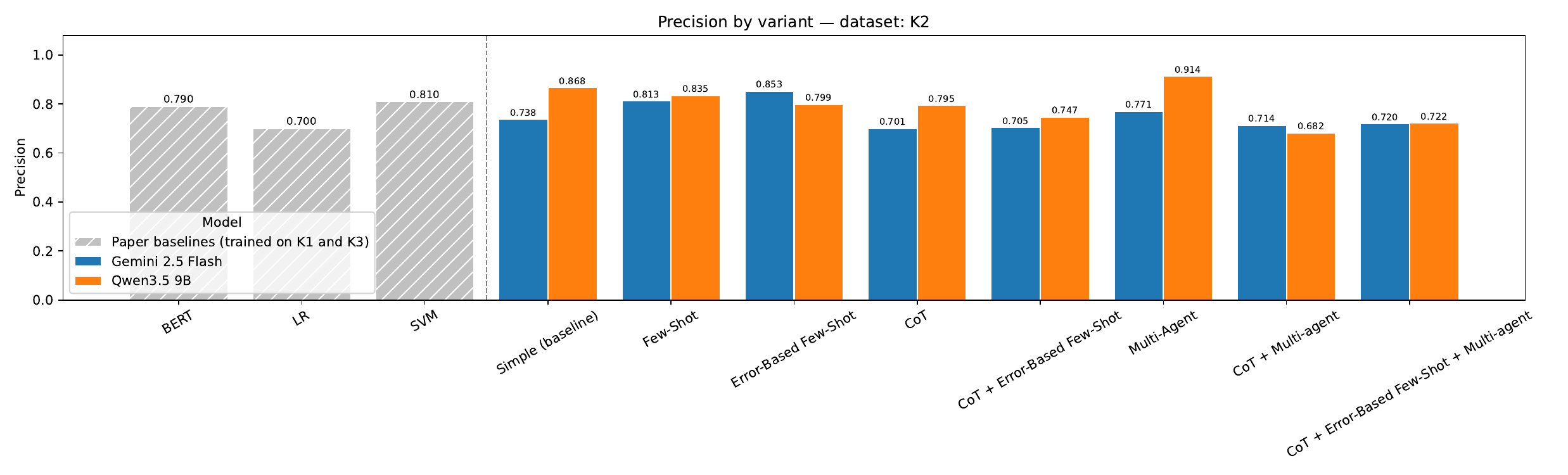}
    \caption{Comparison of precision of deliberative sentences (AD) in batch K2 for the different models and variants.}
    \label{fig:fig2}
\end{figure*}

\section{Results}

\subsection{Classification Results}
\subsubsection{Combined dataset (K1, K2, K3, R4)}

The results of the classification of the combined dataset (K1, K2, K3, R4) can be found in table~\ref{tab:table1}. 
For the local model (Qwen3.5 9B), we observe that the zero-shot baseline achieves similar precision compared to the highest performing model (BERT) from earlier work (0.708 vs. 0.73), while recall is substantially lower (0.458 vs. 0.71). 
This shows that the small model is not very good at predicting deliberativeness out-of-the-box. 
Adding examples (few-shot) already shows improvement in terms of recall (0.603) at the cost of precision (0.58). 
Using only "hard" sentences as examples (Few-Shot Error-Based) again improves classification by both recall (0.615) and precision (0.646). 
Forcing the LLM to reason about whether a sentence is predecisional and deliberative (CoT) results in a jump in recall (0.746). 
Combining CoT with Error-Based few-shot prompting achieves the highest recall (0.8), F1 (0.646) and F2 (0.731) scores of all Qwen3.5 9B variants, and also outperforms the LR, SVM and BERT baselines in recall and F2, but not in precision. 

Then, we move to the multi-agent setups. 
The majority vote variant (Multi-Agent) achieves notable precision (0.769) and beats both the zero-shot baseline and methods from earlier work. 
The high precision, however comes at a high price: it results in the lowest recall (0.4) and F2-score (0.442) of all tested variants. 
Combining multiple agents with Chain-of-Thought reasoning and optionally with error-based few-shot prompting shows minor improvements in recall (respectively 0.477 and 0.456) but precision is now about the same as the best scoring other variants. 
This unexpectedly indicates that a multi-agent setup with a small local model does not seem very good at sensitivity classification. 

In contrast to the local model, the commercial model (Gemini 2.5 Flash) seems to work much better out-of-the-box. 
Surprisingly, the zero-shot baseline even performs better than the other variants on the target measures recall (0.808) and F2 (0.74). 
Few-shot variants show no improvements, the recall and F2 scores are even substantially lower. 
Chain-of-Thought improves over few-shot prompting, and the simplest Multi-Agent setup (majority vote) performs second best, behind the zero-shot baseline. 
These are very different from the results of the local model, where the multi-agent variant performed the worst in terms of recall. 
Various combinations of variants also show slightly lower scores. Where the small Qwen3.5 9B model benefits greatly from more complex variants and combinations when compared to the zero-shot baseline, the commercial model seems already capable of sensitivity classification from its parametric knowledge. 

However, the most important finding of comparing the two models is that the best performing local variant (CoT + Error-Based Few-Shot) very closely approaches the best scoring variant of the commercial model on all measures. 
This is a positive result, as it is a much smaller model which can be run on local infrastructure, which is an important real-world advantage.

\subsubsection{Hard batch (K2)}
Figure~\ref{fig:fig1} shows performance on the K2 subset, which has been identified in prior work as the most "difficult" batch (yielding a relatively low recall score). 
Here, we find major improvements in recall for the best performing variant (CoT + Error-Based Few-Shot) using both the local and commercial models.
Recall almost doubles when compared to SVM (0.754 for the local model and 0.827 for the commmercial model vs. 0.42), and also substantially improves over the better performing BERT (0.57) and LR (0.6) methods, while the precision (in figure \ref{fig:fig2}); already high for the other paper baselines, remains more or less stable. 

Interestingly, for the local model, the Chain-of-Thought and error-based few-shot variants slightly outperform the zero-shot baseline in recall (0.827 vs. 0.796). 
This contrasts with the results on the combined dataset, indicating that bigger models can still benefit from using more complex variants.

\subsection{Analysis of Deliberativeness}

To better understand what type of linguistic features the LLM picks up, we analyze the syntactical differences between sentences that are correctly classified as deliberative, and the sentences correctly classified as non-deliberative.
For this, we create two sets:

\begin{enumerate}
    \item All deliberative (label=1) sentences (N=268) that all eight variants predicted correctly: we call this set \texttt{easy-1}
    \item All non-deliberative (label=0) sentences (N=1,750) that all eight variants predicted correctly: we call this set \texttt{easy-0}
\end{enumerate}

We study these sets \texttt{easy-1} and \texttt{easy-0} to verify whether the LLM has picked up on the linguistic markers of (non-)deliberativeness that are known from literature. 
Since verbs are well-established grammatical markers of activities and actions, we first analyze the use of verbs in both sets of sentences, and report the ten most frequent verbs per set in table \ref{tab:tab-verbs}. 
Reassuringly, we find a strong presence of modal verbs like \emph{'would'}, \emph{'could'} and \emph{'should'} in \texttt{easy-1}, in line with what one would expect from deliberative sentences. 
And, not reported in the table, opinion verbs like \emph{'believe'} (6.7\%), \emph{'propose'} (5.6\%) \emph{'recommend'} (4.8\%) and \emph{'suggest'} (3.4\%) are much more present in \texttt{easy-1} than in \texttt{easy-0}. 

\begin{table}[!h]
    \centering
    \begin{tabular}{lr|lr}
    \toprule
       \textbf{\texttt{easy-0}} & \textbf{\% Sentences} & \textbf{\texttt{easy-1}} & \textbf{\% Sentences}\\
       \midrule
       be & 38.3\% & be & 62.3\% \\
       have & 19.4\% & do & 20.9\% \\
       will & 7.5\% & would & 20.9\% \\
       do & 5.3\% & have & 18.3\% \\
       include & 4.4\% & could & 18.3\% \\
       say & 3.8\% & should & 17.2\% \\
       work & 3.7\% & think & 13.8\% \\
       provide & 3.3\% & will & 13.8\% \\
       make & 3.3\% & want & 11.6\% \\ 
       require & 2.9\% & need & 10.4\% \\
       \bottomrule
    \end{tabular}
    \caption{Occurrences of verbs in \texttt{easy-0} and \texttt{easy-1} sets.}
    \label{tab:tab-verbs}
\end{table}

Building upon these observations, we study verb groups that are commonly used to express opinions: 
\emph{stative verbs} \cite{lakoff1966stative}, 
\emph{reporting verbs} \cite{thompson1991evaluation}, 
\emph{modal verbs} \cite{boyd1969semantics} and 
\emph{cognitive verbs} \cite{fetzer2008and}. 
We rely on the fact that, to qualify as deliberative process privilege, a record must be both \emph{predecisional} and \emph{deliberative}. 
These verb groups are well-known indicators of expressing opinion, and thus match deliberativeness. 
In addition, we augment these with first-person words (\emph{'I'}, \emph{'we'}, \emph{'me'}, ...) as additional indicators, since \emph{personal opinions} are a class of deliberativeness. 
Finally, we study future temporal words, such as \emph{'later'}, \emph{'soon'} and \emph{'tomorrow'}, to validate whether the LLMs picks up the predecisional nature of sentences. 
We also analyze whether the sentence is in past or present tense, the latter potentially indicating that an action has already happened, thus reducing its likelihood of being predecisional. 
Our indicators, alongside some examples can be found in table \ref{tab:indicators}.

\begin{table}[h]
    \centering
    \begin{tabularx}{\linewidth}{cc}
    \toprule
       \textbf{Indicator} & \textbf{Examples} \\
       \midrule
       Stative Verbs & want, prefer, agree, remain  \\
       Reporting Verbs & suggest, indicate, propose, recommend \\
       Modal Verbs & can, could, might, should \\
       Cognitive Verbs & believe, know, understand, assume \\
       First-Person Words & I, we, me, my, our \\
       Future Temporal Words & later, soon, tomorrow, eventually \\
       \bottomrule
    \end{tabularx}
    \caption{Indicators for deliberativeness.}
    \label{tab:indicators}
\end{table}

The proportion of sentences that contain the individual indicators can be seen in Table~\ref{tab:indicators_occur}. 
We observe that verbs groups for expressing opinions (stative, reporting, modal, and cognitive verbs) are much more frequent in deliberative sentences (\texttt{easy-1}) than in non-deliberative sentences (\texttt{easy-0}). 
With respect to the temporal dimension, future temporal markers are generally not frequent, although they do occur twice as often in deliberative sentences as in non-deliberative ones. 
Non-deliberative sentences are more often written in past tense, but the difference is minor. 

\begin{table*}[h]
    \centering
    \begin{tabular}{cccccccc}
        \toprule
       \textbf{Set} & \textbf{Stative} & \textbf{Reporting} & \textbf{Modal} & \textbf{Cognitive} & \textbf{First-Person} & \textbf{Future Temporal} & \textbf{Past Tense} \\ 
       \midrule
       easy-0 & 55.4\% & 17.4\% & 16.9\% & 6.1\% & 35.2\% & 1.7\% & 53.1\%  \\
       easy-1 & 82.5\% & 29.5\% & 75.0\% & 32.1\% & 84.7\% & 3.4\% & 47\%  \\
       % \midrule
       % TP->FN & 77.1\% & 27.1\% & 59.9\% & 15.8\% & 42.6\% & 4\% & 48.3\% \\
       % FN->TP & 71.3\% & 32.6\% & 39.5\% & 15.5\% & 39.5\% & 3.1\% & 60.5\% \\
       % \midrule
       % TN->FP & 56.8\% & 21.1\% & 34.2\% & 7.8\% & 21.1\% & 2.2\% & 46.9\% \\
       % FP->TN & 62.7\% & 18.3\% & 48.9\% & 12.9\% & 30.8\% & 2.3\% & 39.8\% \\
       \bottomrule
    \end{tabular}
    \caption{Percentages of sentences in which a specific indicator occurs.}
    \label{tab:indicators_occur}
\end{table*}

\begin{table*}[h]
    \centering
    \begin{tabular}{ccccc}
    \toprule
       \textbf{Set} & \textbf{Median unique indicators} & \textbf{Median total indicators} & \textbf{>=2 indicators} & \textbf{0 indicators} \\
       \midrule
       easy-0 & 1 & 1 & 29.7\% & 31.0\%  \\
       easy-1 & 3 & 4 & 92.2\% & 0.4\%  \\
       % TP->FN & 2 & 3 & 73.3\% & 4.6\% \\
       % TN->FP & 1 & 2 & 44.1\% & 24.8\% \\
       % FN->TP & 2 & 3 & 64.3\% & 10.9\% \\
       % FP->TN & 2 & 2 & 57.9\% & 15.9\% \\
       \bottomrule
    \end{tabular}
    \caption{Analysis of co-occurrences of indicators.}
    \label{tab:indicators-cooccur}
\end{table*}

\begin{table*}[h]
    \centering
    \begin{tabular}{ccp{0.7\linewidth}}
    \toprule
       \textbf{\#Indicators} & \textbf{Label} & \textbf{Example} \\
       \midrule
       0 & 0 & The Welfare to Work Tax Credit, enacted in the 1997 Balanced Budget Act, provides a credit equal to 35 percent of the first \$10,000 in wages in the first year of employment, and 50 percent of the first \$10,000 in wages in the second year, to encourage the hiring and retention of long term welfare recipients \\
       0 & 1 & Give all primary civil authority to HCFA, but establish a referral process in an MOU for both chains and individual facilities. \\
       \midrule
       1 & 0 & Below is a brief summary of the other important items that were \textbf{discussed}. \\
       1 & 1 & OMB \textbf{suggested} increasing the local match rate over time \\
       \midrule
       2 & 0 & Q What \textbf{will} the overall increase in funding \textbf{be} for drug testing and treatment as a result of this new initiative? \\
       2 & 1 & \textbf{We} \textbf{could} also reintroduce a version of education opportunity zones. \\
       \midrule
       3 & 0 & A number of states, including Louisiana and Wisconsin, financial officer, whom prosecutors \textbf{have} accused of orchestrating a have passed laws \textbf{saying} that organs donated in the \textbf{state} can't \textbf{be} massive fraud. \\
       3 & 1 & \textbf{We} \textbf{should} \textbf{be} honest with President about the serious issues at stake here -- especially since 3 of his toughest cabinet members are not in agreement. \\
       \midrule
       4 & 0 & Let us \textbf{remember} the difficult years chronicled in this report, and \textbf{think} about how good people \textbf{could} have done things that \textbf{we} \textbf{know} were wrong. \\
       4 & 1 & However, \textbf{I} don't \textbf{think} that the policy determination can \textbf{be} completely divorced from an \textbf{assessment} of the political overlay on this issue. \\
       \midrule
       5 & 0 & Speaker Bustamante \textbf{says} that \textbf{we} \textbf{will} \textbf{know} in the next two weeks whether there is \textbf{agreement} on the Cal. bilingual  education legislation. \\
       5 & 1 & \textbf{I} don't \textbf{know} where things stand in the negotiations over this amendment, but it \textbf{would} \textbf{be} great if we \textbf{could} \textbf{indicate} that this particular provision \textbf{would} \textbf{be} a deal breaker. \\
       \bottomrule
    \end{tabular}
    \caption{Example sentences with their label and number of unique co-occurrences of indicators}
    \label{tab:example_sentences}
\end{table*}

In table \ref{tab:indicators-cooccur} we report the co-occurrences of indicators (e.g., when a stative and reporting verb occur together in a single sentence), excluding the past tense indicator as it was not sufficiently discriminative. 
Deliberative sentences often contain three unique indicators, and four total indicators (a single indicator category can occur multiple times). 
On the other hand, non-deliberative sentences mostly contain only one. 
Almost all deliberative sentences contain more than one (>=2) unique indicators, and practically none have no indicators at all. 
This is markedly different for non-deliberative sentences, where only about a third of the sentences contain multiple unique indicators, and another third has no indicators. 
We found that for deliberative sentences, a combination of \texttt{First-Person + \{Stative, Reporting, Modal, Cognitive\}} was most frequent. 
Lastly, in Table~\ref{tab:example_sentences}, we show examples of sentences with the number of indicators ranging from zero to five (no sentences contained all six), with indicators highlighted in bold. 
It shows that deliberative sentences contain a high number of indicator categories, which clearly represent opinions, while non-deliberative sentences with a high number of indicator categories that contain verb groups known for expressing opinions (such as \emph{'will'} or \emph{'could'}, are more frequently in third-person, clearly representing something else than a personal opinion.

\section{Discussion}
We find that the difference in recall between single-agent and multi-agent setups is the most notable difference. 
We ran a small analysis over the results where the first (predictor) and second (critic) agents disagree, with the third (judge) agent following the critic (N=1,538). 
These results show that more (476) sentences were incorrectly changed from deliberative to non-deliberative, while only 129 sentences were correctly changed from non-deliberative to deliberative, thus hurting recall. 
Overall, the critic and judge agents seem to be more conservative in predicting sentences as deliberative, as 1,087 were corrected to non-deliberative, but only 451 sentences were corrected to deliberative. 
Also, they seem to be good signalers of false positives (611 times), but too eager, resulting in also a lot of true positives being flagged as false positives (476 times), which especially hurts recall. 

Although we use the expert reviewer annotations from \citet{branting2025decision}, sensitivity classification is an inherently subjective task and the FOIA law texts leave room for different interpretations. 
The used dataset contains only documents of two custodians and all batches are annotated by the authors of the paper. 
Future work should investigate the performance and generalizability of this approach to a more diverse set of documents. 
However, our approach already has the advantage that the classification model requires no training, which partly solves the problem. 

Furthermore, the dataset is deliberately annotated to judge solely on the sentence itself, irrespective of the broader context of the document. 
This differs from the real-world situation, where sensitivity reviewers make use of the context of the document when deciding to redact a sentence or not. 
However, as the authors of the dataset correctly argue, these new annotations make this a more tractable text classification task. 
It would be interesting to see whether supplying more context to the language model improves classification. 
This would be an extension of the older dataset from \citet{baron2022providing}, which took into account the surrounding context, but was only annotated on the paragraph level. 
This is a harder task, but the context may also bring new opportunities for improving classification with LLMs.

Finally, we use Gemini 2.5 Flash and Qwen3.5 9B as LLMs, but using other models may very well produce different results. 
The goal of this paper was not to thoroughly compare models, but rather assess how well a state-of-the-art small, local model was able to make predictions on the sensitivity of sentences, as compared to a realistic alternative: a frequently used, cost-effective and fast commercial closed model. 

\section{Conclusion}
In this work, we systematically evaluated whether a small, local LLM (Qwen3.5 9B) is able to automatically detect sensitive sentences according to the FOIA Exemption 5 deliberative process privilege in government documents. 
We compared eight well-known prompting techniques for improving text classification with LLMs: zero-shot, Chain-of-Thought reasoning, few-shot examples, multi-agent and various combinations thereof. 

Using a collection of White House government documents, with annotations for deliberativeness on the sentence-level, we show that combining Chain-of-Thought reasoning with error-based few-shot examples achieves the highest recall and F2 score. 
It substantially outperformed the Zero-Shot baseline, as well as LR, SVM and BERT baselines from earlier work. 
Also, this variant very closely approaches the best scoring variant using Gemini 2.5 Flash, a popular commercial closed model. 

We also performed an analysis based on the classification results to validate the syntactical indicators of deliberativeness when comparing obviously deliberative sentences with obviously non-deliberative sentences (all variants agree on classification). 
Results show that sentences classified as deliberative contain a higher number of verbs commonly used for expressing opinions (stative, reporting, modal, cognitive verbs) and are more often in first-person. 
Sentences classified as deliberative are also characterized by the presence of multiple indicators at the same time, most notably combinations of first-person words with verbs commonly used for expressing opinions. 
For future work, it would be interesting to see whether these indicators can be used to aid LLMs in making sensitivity predictions.

\begin{acks}
Maik Larooij and David Graus are funded by \grantsponsor{ICAI}{ICAI (AI for Open Government Lab)}{https://icai.ai/}. 
Views expressed in this paper are not necessarily shared or
endorsed by those funding the research.
\end{acks}

\bibliographystyle{ACM-Reference-Format}
\bibliography{samples/sample-base}

\end{document}